\documentclass{Interspeech2024}

\usepackage[capitalize]{cleveref}
\usepackage{multirow}

\interspeechcameraready 

\title{Quantifying the Role of Textual Predictability in Automatic Speech Recognition}

\name{Sean}{Robertson}
\name{Gerald}{Penn}
\name{Ewan}{Dunbar}

\address{University of Toronto, Canada}
\email{sdrobert@cs.toronto.edu,gpenn@cs.toronto.edu,ewan.dunbar@utoronto.ca}

\keywords{speech recognition, perplexity, entropy, language model, acoustic model, accent-robust speech recognition, African American English}

\begin{document}

\maketitle

\begin{abstract}
A long-standing question in automatic speech recognition research is how to attribute errors to the ability of a model to model the acoustics, versus its ability to leverage higher-order context (lexicon, morphology, syntax, semantics). We validate a novel approach which models error rates as a function of relative textual predictability, and yields a single number, $k$, which measures the effect of textual predictability on the recognizer. We use this method to demonstrate that a Wav2Vec 2.0-based model makes greater stronger use of textual context than a hybrid ASR model, in spite of not using an explicit language model, and also use it to shed light on recent results demonstrating poor performance of standard ASR systems on African-American English. We demonstrate that these mostly represent failures of  acoustic--phonetic modelling. We show how this approach can be used straightforwardly in diagnosing and improving ASR.
\end{abstract}

\section{Introduction} \label{sec:intro}

Recent work has highlighted the difficulties  automatic speech recognition (ASR) systems continue to have with minority and racialized language varieties. However, while all studies agree that the  ultimate source of the problem is the change in domain---ASR training is generally on dominant language varieties---explanations for issues with African-American English in particular vary, with some  arguing that many issues stem from  morphological and vocabulary differences \citep{martinUnderstandingRacialDisparities2020}, while others that phonetic differences are the main source \citep{koeneckeRacialDisparitiesAutomated2020,wassinkUnevenSuccessAutomatic2022}. These questions put into relief a long-standing question in ASR: how to assess how much a system relies on textual predictability (traditional ``language modelling'') versus modelling of the phonetic signal (traditional ``acoustic modelling''). This has become  difficult to resolve with the advent of powerful end-to-end models deploying context at long distances, reducing or eliminating the need for explicit language models.

We develop a new method for quantifying the role of textual predictability in ASR, starting from a psychoacoustic paradigm developed by \citet{boothroydMathematicalTreatmentContext1988}. We validate the use of this framework for automatic (as opposed to human) speech recognition by demonstrating  that  utterances with different degrees of textual predictability  yield increasing values of $k$. We also show that a more powerful explicit language model yields higher values of $k$, indicating stronger textual prediction.

We apply the method to comparing ASR models that we expect to have different intrinsic capacities for contextual predictability (GMM,  TDNN, Wav2Vec 2.0-base, and Wav2Vec 2.0-large), demonstrating that $k$ also increases with more powerful models. We also apply the method to an African-American English corpus \citep{kendallCorpusRegionalAfrican2023}, reaching a similar conclusion to previous works \citep{koeneckeRacialDisparitiesAutomated2020,wassinkUnevenSuccessAutomatic2022}: the difficulties faced by ASR systems with these language varieties mainly reflect issues with acoustic modelling. We provide a recipe for using this method to diagnose issues and improve performance in ASR, and discuss its limitations. All of our code and results are open source and available at
\ifinterspeechfinal%
\url{https://github.com/sdrobert/kaldi-boothroyd}%
\else%
[to ensure author anonymity, the link to the resource will be added after the review process]%
\fi.

\section{Background} \label{sec:related}


\subsection{ASR and textual predictability} \label{sec:related_asr}

Textual predictability in ASR is typically measured using perplexity as measured by some language model (LM). For the distribution $Q$ induced by an LM, perplexity is the exponent of the negative log likelihood (NLL) $H_y$ of a token sequence $y = y_1, y_2, \ldots, y_L$, formally:
\begin{equation}
    H_y = -\frac{1}{L} \log Q(y). \label{eq:nll}
\end{equation}
\Cref{eq:nll} is an estimate of the cross-entropy rate $\mathbb{E}_y[H_y]$ of $Q$ relative to the population distribution $P$ which generates $y$ \citep{manningFoundationsStatisticalNatural1999}. Since a lower $H_y$ implies a higher $Q(y)$, the NLL measures how well $Q$ predicts $y$, and, if we average $H_y$ over a corpus drawn from $P$, how well $Q$ predicts $P$.

We expect NLL calculated with respect to $Q$ to be correlated with ASR accuracy: whether the ASR system uses an explicit language model following $Q$ or not, assuming that the system is trained on data following $P$, the system has an implicit marginal textual distribution $Q_{sys}$: for a transcription $y$, and where $x \in \mathcal{X}$ is the set of all possible utterances:
\begin{equation}
Q_{sys}(y) = \sum_{x\in\mathcal{X}} Q_{sys}(y|x)P(x)
\end{equation}
Because of the shared training data, we expect $Q_{sys}$ to be fairly close to both $P$ and to some LM distribution $Q$.

Indeed, NLL was proposed as a measure of the intrinsic difficulty of transcribing an utterance \citep{bahlMaximumLikelihoodApproach1983}, with some attempts at modelling the relationship between ASR error rates $e_y$, $0 \leq e_y \leq 1$, and $H_y$  \citep{printzTheoryPracticeAcoustic2002,klakowTestingCorrelationWord2002,chenEvaluationMetricsLanguage2008}. \citet{klakowTestingCorrelationWord2002} suggest the following power law relationship with fit coefficients $a,b \in \mathbb{R}$, that is, log error rates being proportional to $H_y$:
\begin{equation}
    e_y = b\exp(a H_y). \label{eq:klakow}
\end{equation}
\Cref{eq:klakow} would be a strong candidate for quantifying the role of textual predictability on ASR performance were it not sensitive to ``acoustic conditions.'' As remarked by \citet{klakowTestingCorrelationWord2002}, the coefficient $a$ decreases (while $b$ grows) as acoustic conditions become more ``challenging.'' Thus, \cref{eq:klakow} is unlikely to generalize across corpora.
Rather than attempt to link NLL directly to performance, we propose to work using ratios, relating \emph{relative} predictability to \emph{relative} performance. Furthermore, we construct a measure which is aggregated over acoustic conditions of increasing difficulty, in an attempt to further factor out the role of acoustics.

\subsection{Predictability and performance} \label{sec:related_rel}

\begin{figure}
    \centering
    \includegraphics{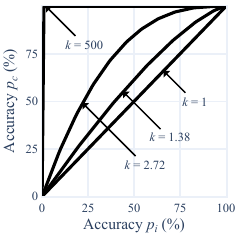}
    \caption{Accuracy ratios across $k$ from Equation \eqref{eq:k}.} \label{fig:bn}
\end{figure}

Our method is based on  the experimental paradigm of \citet{boothroydMathematicalTreatmentContext1988}, in which  participants recognized sentences across three conditions: zero predictability (\textbf{ZP})---words drawn randomly---low predictability (\textbf{LP})---grammatical but semantically strange---and high predictability (\textbf{HP}). Error rates $e$ and accuracies $p = 1 - e$ were computed per condition, inducing errors by masking the speech over a range of signal-to-noise ratios (SNRs). Treating \textbf{ZP} as the ``isolated'' condition $i$ and either \textbf{LP} or \textbf{HP} as the ``context'' condition $c$, the authors found that error rates were related by a constant exponent $k$, regardless of the SNR range:
\begin{equation}
    e_c = e_i^k,\text{ or }p_c = 1 - (1 - p_i)^k \label{eq:k}
\end{equation}
Figure \ref{fig:bn} illustrates the relation between $p_c$, $p_i$, and $k$, where  variation in accuracy is induced by varying SNR. $k = 1$ means the listener is not using the additional predictability of condition $c$ to compensate for acoustics, whereas for $k = 500$, the listener leverages so much context as to make  acoustics irrelevant. In  \cite{boothroydMathematicalTreatmentContext1988}, a greater gap in predictability  led to greater $k$: between \textbf{ZP}--\textbf{LP}, $k \approx 1.38$, and between \textbf{ZP}--\textbf{HP}, $k \approx 2.72$. 
The  design is easily transposed to ASR. 
While the result that $k$ is independent of SNR range has not always held up with human listeners \cite{nittrouerContextEffectsPhoneme1990,grantRecognitionIsolatedWords2000,bronkhorstModelContextEffects1993,bronkhorstEvaluationContextEffects2002}, we show it is a useful approximation for ASR. 

Our method is as follows. Given an evaluation corpus, we split it into bins by textual predictability. We bin using the NLL of a LM trained on the same distribution as the target system. Call the training distribution $P_{train}$, the binning LM distribution $Q_{bin}$, and the marginal textual distribution of the target system $Q_{sys}$. We divide into three (or more) bins: one reference ($i$) bin (for which we continue the misnomer \textbf{ZP}), and two bins, \textbf{LP} (more predictable than \textbf{ZP}) and \textbf{HP} (much more predictable).

On in-domain data, we expect higher $k$ when $c$ is \textbf{HP} than when it is \textbf{LP}. Systems that, intuitively, ``rely more on language modelling'' are those for which $Q_{sys}$ is closer to $P_{train}$. We expect a pronounced gap between \textbf{HP} and \textbf{LP} for these systems.

For evaluation data following an unknown distribution $P_{test}$, we keep the same NLL cuts, and continue to use $Q_{bin}$ (trained on $P_{train}$). We consider two different cases. If the percentage of the new corpus in each bin is very different than an in-domain corpus, this suggests that $Q_{bin}$ is severely mismatched to $P_{test}$. We can assume that $Q_{sys}$ is also mismatched. The $k$ calculated on in-domain data then tells us how sensitive the system should be to this textual domain shift: $k=1$ should be insensitive, extremely high $k$ should be catastrophic. On the other hand, it is possible that the bin frequencies reveal no major mismatch between $Q_{bin}$ and $P_{test}$. Since we are extrapolating to $P_{test}$, this provides no guarantee that the ASR system is well-matched. However, if $Q_{sys}$ is a poor match to $P_{test}$, we predict that $k$ should be lower on out-of-domain than in-domain data, approaching 1.

\section{Experiments} \label{sec:experiments}

\subsection{Materials and systems} \label{sec:experiments_materials}

We experiment with LMs and ASR systems from Kaldi's \emph{s5} recipe \citep{poveyKaldiSpeechRecognition2011} and Wav2Vec 2.0 \citep{baevskiWav2vecFrameworkSelfsupervised2020}. Kaldi LMs\footnote{
    Available at \url{https://kaldi-asr.org/models/m13} and \url{https://openslr.org/11/}, last accessed February 19, 2024.%
} include: a pruned, word-level, 3-gram LM with modified Kneser-Ney smoothing \citep{chenEvaluationMetricsLanguage2008}; similarly, an un-pruned 4-gram and a word-level, recurrent neural network (RNN) \citep{xuNeuralNetworkLanguage2018}. Kaldi acoustic models include a speaker adaptive Gaussian mixture model (GMM) and a time-delay neural network (TDNN).\footnote{%
    The TDNN is available at \url{https://kaldi-asr.org/models/m13}, last accessed February 19, 2024. We have uploaded our re-trained GMM, denoted \emph{tri6b} in the Kaldi s5 recipe, to our repository.%
} We denote  GMM and TDNN ASR systems with lattices weighted by the 3-gram LM as \textbf{GMM-3} and \textbf{TDNN-3}, respectively. We denote the ASR system which re-scores \textbf{TDNN-3} lattices with the 4-gram LM as \textbf{TDNN-4}. We  also apply two fine-tuned Wav2Vec 2.0 models available freely online. The ``base'' variant (\textbf{W2V2-B}), features 12, smaller Transformer layers and is trained on LibriSpeech \citep{panayotovLibrispeechASRCorpus2015}.\footnote{%
    Available at \url{https://huggingface.co/facebook/wav2vec2-base-960h}, last accessed February 19, 2024.%
} The ``large'' variant, denoted \textbf{W2V2-L}, has 24, larger Transformer layers and has been additionally trained on LibriLight  \citep{kahnLibrilightBenchmarkASR2020}. Both have been fine-tuned for ASR on LibriSpeech with a CTC objective \citep{gravesConnectionistTemporalClassification2006}. We use greedy decoding without an external LM for simplicity.
The  
hybrid models offer explicit control of the amount of language modelling the ASR system is doing, speaking directly to our hypotheses. Wav2Vec 2.0 allows us to explore the role of implicit textual prediction: since these networks use global self-attention \citep{vaswaniAttentionAllYou2017}, we expect them to use predictive context more aggressively.

Within-domain, we compute error rates and fit $k$ values on LibriSpeech's \emph{dev-clean} and \emph{dev-other} partitions (\textbf{LS-C} and \textbf{LS-O} respectively). Following prior work \cite{martinUnderstandingRacialDisparities2020,koeneckeRacialDisparitiesAutomated2020}, we expect these systems to under-perform on utterances from the Corpus of Regional African American Language (CORAAL) \citep[version~2023.06]{kendallCorpusRegionalAfrican2023}. In particular, we focus on the utterances of speakers from Rochester, New York (\textbf{CL-R}) and Princeville, North Carolina  (\textbf{CL-P}), on which \cite{koeneckeRacialDisparitiesAutomated2020} reported the  lowest ($e_y = 0.20$) and highest ($e_y = 0.38$) error rates from the corpus, respectively.\footnote{
    Compared to \citet{koeneckeRacialDisparitiesAutomated2020}, we sanitize the partitions more aggressively to more closely resemble a standard ASR benchmark: any utterances containing restarts, fillers, unintelligible markers, non-speech noise, and so forth are excluded from consideration. After filtering, the CL-R and CL-P partitions contain roughly 4 and 3 hours of speech, respectively. Filtering is reproducible from our code base.%
} 

\subsection{Procedure} \label{sec:experiments_procedure}

Utterances are  first corrupted by noise over a range of SNRs, and decoded by each ASR system, on each corpus partition. We follow the procedure of \citet{zhangEstimateNoiseEffect2023} for introducing noise to utterances. Each recording is first normalized to a fixed reference power and 0 DC. Then, white noise of SNRs between -10 and 30 dB is added to each recording. As \citeauthor{zhangEstimateNoiseEffect2023} found that different types of generated noise lead to similar accuracies at similar SNRs, we did not experiment with different types of noise. For the sake of our analysis, it is sufficient that noise degrades acoustic conditions consistently across NLL bins.

For binning,  following \cref{sec:related_asr}, an LM with a low NLL is considered close to the training distribution $P_{train}$. We used the RNN LM to generate the bins as it produced the lowest NLL on \textbf{LS-C} and \textbf{LS-O}. 
The cutpoints (same across corpora) were obtained by evenly splitting per-utterance NLL from  \textbf{LS-C}  into three intervals. Because the tails of the distribution were long, we dropped the top and bottom 5\% of NLLs before constructing the bins. The \textbf{HP} bin covers $H_y \in (3.4,4.5]$, the \textbf{LP} bin $H_y \in (4.5, 5.6]$, and the \textbf{ZP} bin ($i$ condition) $H_y \in (5.6, 6.8]$.

To estimate $k$, we calculate a single error rate $e$ per system/corpus/bin triplet. We take the \textbf{ZP} rate $e_i$ as our ``isolated'' or ``no-context'' condition and either the \textbf{LP} or \textbf{HP} rate $e_c$ as our ``context'' condition, fitting $k$ to \cref{eq:k}.
We perform non-linear least-squares regression to fit $e_c = e_i^k$. We could perform ordinary least squares on $\ln e_c = k \ln e_i$ instead, but because the residuals are smaller when $e_c \approx 1$, we found this biased the fit to the lowest values of $e_c$. To compute 95\% confidence intervals for each fit of $k$, we rely on the Wild bootstrap \citep{wuJackknifeBootstrapOther1986}: for each of $B = 9999$ iterations, we resample the log-space residuals $\widehat{\epsilon} = \epsilon V$, where $V \sim \mathcal{N}(0, 1)$, re-compute $\ln \widehat{e}_c = k \ln e_i + \widehat{\epsilon}$, and re-fit $\widehat{k}$. The log-space ensures $\widehat{e}_c > 0$; multiplication $\epsilon V$ maintains heteroskedasticity of the residuals.


\section{Results} \label{sec:results}


\begin{table}
\centering
\caption{%
    Word error rates $e_y$, reported as a percentage. Rows are grouped by partition and NLL bin; columns by model. The \emph{all} rows contain the error rates over the entire partition, without binning.
} \label{tab:wer}
\footnotesize
\begin{tabular}{>{\raggedright}p{1.2em}>{\raggedright}p{1.1em}|*{5}{c}}
\toprule
 & & \scriptsize GMM-3 & \scriptsize TDNN-3 & \scriptsize TDNN-4 & \scriptsize W2V2-B & \scriptsize W2V2-L \\

\midrule
\multirow[t]{4}{*}{LS-C} & HP & 8.4 & 3.7 & 2.4 & 2.2 & \bf 1.5 \\
 & LP & 11.1 & 4.9 & 3.5 & 3.3 & \bf 2.2 \\
 & ZP & 16.2 & 7.8 & 5.9 & 6.6 & \bf 4.4 \\
\cline{2-7}
 & all & 10.5 & 4.7 & 3.3 & 3.3 & \bf 2.2 \\
\cline{1-7}
\multirow[t]{4}{*}{LS-O} & HP & 22.1 & 10.0 & 6.5 & 6.3 & \bf 3.2 \\
 & LP & 28.4 & 13.1 & 9.7 & 10.0 & \bf 5.2 \\
 & ZP & 37.0 & 18.7 & 15.3 & 16.2 & \bf 8.5 \\
 \cline{2-7}
 & all & 26.1 & 12.2 & 8.7 & 8.8 & \bf 4.6 \\
 \cline{1-7}
\multirow[t]{4}{*}{CL-R} & HP & 45.8 & 31.2 & 26.4 & 25.2 & \bf 14.6 \\
 & LP & 54.0 & 38.5 & 35.2 & 31.5 & \bf 20.8 \\
 & ZP & 58.1 & 43.6 & 41.6 & 38.2 & \bf 26.0 \\
 \cline{2-7}
 & all & 53.9 & 37.3 & 33.9 & 32.8 & \bf 23.2 \\
\cline{1-7}
\multirow[t]{4}{*}{CL-P} & HP & 73.5 & 56.2 & 52.4 & 50.6 & \bf 37.8 \\
 & LP & 79.2 & 64.1 & 62.1 & 59.2 & \bf 47.0 \\
 & ZP & 83.3 & 69.0 & 68.6 & 66.0 & \bf 54.3 \\
 \cline{2-7}
 & all & 78.7 & 61.9 & 59.4 & 58.2 & \bf 46.2 \\

\bottomrule
\end{tabular}
\end{table}

Table \ref{tab:wer} lists  average error rates,  without noise. In general, \textbf{GMM-3} has the most errors, then \textbf{TDNN-3}, \textbf{TDNN-4}, \textbf{W2V2-B}, and \textbf{W2V2-L} the fewest. Further, as expected \citep{bahlMaximumLikelihoodApproach1983,klakowTestingCorrelationWord2002}, error rates increase as a function of NLL. Finally, we note wide disparity between  LibriSpeech  and  CORAAL.

\begin{table}
    \centering
    \caption{%
    Estimated $k$  and bootstrapped 95\% confidence intervals. The first block lists fit $k$ values per partition, averaged over models. The second block is per model, averaged over partitions. The \emph{all} row aggregates all models per partition.} \label{tab:k}
\footnotesize
\begin{tabular}{rr|cccc}
\toprule
 &  & \multicolumn{2}{c}{HP} & \multicolumn{2}{c}{LP} \\

& & $k$ & CI  & $k$ & CI \\
\midrule
\multirow[t]{6}{*}{LS-C} & GMM-3 & 1.34 & { [1.33, 1.36]} & 1.21 & { [1.20, 1.22]} \\
 & TDNN-3 & 1.31 & { [1.30, 1.33]} &  1.19 & { [1.18, 1.20]} \\
 & TDNN-4 & 1.42 & { [1.40, 1.43]} & 1.23 & { [1.22, 1.24]} \\
 & W2V2-B & 1.50&  { [1.45, 1.56]} & 1.19 & { [1.17, 1.20]} \\
 & W2V2-L & 1.57 & { [1.53, 1.62]} & 1.16&  { [1.14, 1.17]} \\
 \cline{2-6}
 & all & 1.40&  { [1.38, 1.42]} & 1.20&  { [1.19, 1.20]} \\
\cline{1-6}
\multirow[t]{6}{*}{LS-O} & GMM-3 & 1.43 & { [1.42, 1.44]} & 1.24& { [1.23, 1.25]} \\
 & TDNN-3 & 1.33 & { [1.31, 1.34]} & 1.17 & { [1.16, 1.19]} \\
 & TDNN-4 & 1.42 & { [1.41, 1.43]} & 1.21&  { [1.20, 1.22]} \\
 & W2V2-B & 1.44 & { [1.41, 1.47]} & 1.19 & { [1.17, 1.20]} \\
 & W2V2-L & 1.47 & { [1.43, 1.50]} & 1.16&  { [1.15, 1.18]} \\
 \cline{2-6}
 & all & 1.41 & { [1.40, 1.42]} & 1.20 & { [1.19, 1.20]} \\
 \cline{1-6}
 \multirow[t]{6}{*}{CL-R} & GMM-3 & 1.50 & { [1.48, 1.52]} & 1.19&  { [1.18, 1.20]} \\
 & TDNN-3 & 1.44 & { [1.42, 1.46]} & 1.17&  { [1.17, 1.18]} \\
 & TDNN-4 & 1.61&  { [1.58, 1.63]} & 1.23&  { [1.22, 1.24]} \\
 & W2V2-B & 1.52 & { [1.50, 1.54]} & 1.23&  { [1.23, 1.24]} \\
 & W2V2-L & 1.56& { [1.51, 1.61]} & 1.24 & { [1.22, 1.25]} \\
 \cline{2-6}
 & all & 1.53 & { [1.51, 1.54]} & 1.22 & { [1.21, 1.22]} \\
\cline{1-6}
\multirow[t]{6}{*}{CL-P} & GMM-3 & 1.66 & { [1.65, 1.68]} & 1.28&  { [1.26, 1.29]} \\
 & TDNN-3 & 1.58 & { [1.57, 1.59]} & 1.22 & { [1.22, 1.23]} \\
 & TDNN-4 & 1.71&  { [1.69, 1.74]} & 1.26&  { [1.24, 1.28]} \\
 & W2V2-B & 1.68&  { [1.66, 1.71]} & 1.26&  { [1.24, 1.27]} \\
 & W2V2-L & 1.70&  { [1.68, 1.73]} & 1.27 & { [1.26, 1.29]} \\
 \cline{2-6}
 & all & 1.67 & { [1.66, 1.69]} & 1.26&  { [1.25, 1.26]} \\
\bottomrule
\end{tabular}
\end{table}

\Cref{tab:k} lists  $k$   by partition, model, and in aggregate. Columns represent the choice of ``context'' bin $e_c$, which is either \textbf{LP} or \textbf{HP}; the ``isolated'' bin is always \textbf{ZP}.   We concentrate on in-domain (\textbf{LS}) first. In all cells, $k > 1$, and confidence intervals do not include $k = 1$: textual predictability plays a role in error rates. Furthermore, $k$ is higher when fit to the \textbf{HP} error rates than the \textbf{LP} rates---$k$ increases as a function of predictability. Finally, on \textbf{HP}, there is a divide between  models using a 3-gram LM versus the more sophisticated models, particularly \textbf{W2V2-L}, with the latter yielding higher $k$ values.

\begin{table}
    \centering
    \caption{Proportion of partition captured by each NLL bin (\%). The \emph{total} column sums each row.} \label{tab:prop}
    \footnotesize
    \begin{tabular}{r|ccc|c}
    \toprule
        & HP & LP & ZP & total \\
    \midrule
LS-C & 37.2 & 40.1 & 12.6 & 89.9 \\
LS-O & 39.9 & 39.5 & 11.2 & 90.6 \\
\hline
CL-R & 18.3 & 38.7 & 26.9 & 84.0 \\    
CL-P & 23.8 & 41.9 & 23.4 & 89.1 \\

    \bottomrule
    \end{tabular}
\end{table}

Next we consider the out-of-domain (\textbf{CL}) data. First, \Cref{tab:prop} tabulates the proportion of utterances per partition captured in each bin. 
The two \textbf{LS} partitions have similar proportions in each bin. On CORAAL, the vast majority of the data remain in the \textbf{HP} and \textbf{LP} bins, in line with the observation of \citet{koeneckeRacialDisparitiesAutomated2020} that CORAAL and LibriSpeech transcriptions are more similar than different. Nevertheless, the mass shifts toward the \textbf{ZP} bin, raising the possibility that CORAAL error rates could be affected by textual predictions, as per \citet{martinUnderstandingRacialDisparities2020}. In the absence of previous studies, it is difficult to say how big such an effect could be: while in-domain $k$ values are greater than $1$, they are far from the catastrophic $k=500$ case. Returning, then, to \Cref{tab:k}, we recall that, in cases of mismatch between $Q_{sys}$ and the new domain $P_{test}$, we expect $k$ to go down, approaching $1$. In fact, in general, $k$ values are \emph{higher} on  CORAAL. Thus,  we reach a similar conclusion to \citet{koeneckeRacialDisparitiesAutomated2020}: when applying (these) ASR systems to African-American English, the effect of textual domain shift is limited---at least as measured by our approach. Further research should be done to explore the relation between in-domain $k$ and out-of-domain performance.


\begin{figure}
    \centering
    \includegraphics{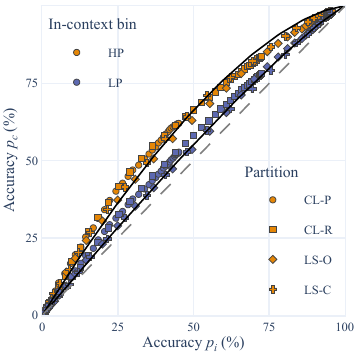}
    \caption{%
        In-context \emph{vs.} isolated accuracies of \textbf{W2V2-L}. The grey, dashed line is $y=x$. Black lines mark the interpolated fits over \textbf{LS-C} from \cref{tab:k}: the shallow curve is \textbf{LP}; the steep curve is \textbf{HP}.
    } \label{fig:acc-ratio}
\end{figure}

As discussed above, we know that, for human listeners, $k$ as calculated using \Cref{eq:k}, does not give a perfect fit to the data. \Cref{fig:acc-ratio} plots accuracy in the isolated  versus the context condition at a fixed SNR on \textbf{W2V2-L}.\footnote{%
    Analogous plots to \cref{fig:acc-ratio,fig:pointwise-k} for other models are included in the supplementary material.
} Colour and shape distinguish context bin and data partition, respectively. Black lines mark the fit $k$ to \cref{eq:k} on \textbf{LS-C}. Data from all partitions follow a similar curve. The fitted $k$ is in broad agreement with this curve, though it overestimates at low isolated accuracies and underestimates at high isolated accuracies.

\begin{figure}
    \centering
    \includegraphics{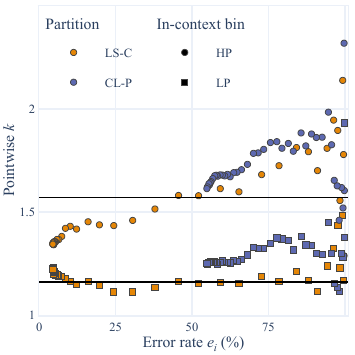}
    \caption{%
        Point-wise estimates of $k = \ln e_c / \ln e_i$ \emph{vs.} error rates $e_i$ of \textbf{W2V2-L}. Each point is paired by SNR and partition. Black lines mark the interpolated fits from \cref{tab:k}.%
    } \label{fig:pointwise-k}
\end{figure}

To better illustrate this, in \Cref{fig:pointwise-k} we compare the point-wise estimates of $k$ versus isolated error rates $e_i$ on \textbf{W2V2-L}. The two fitted values of $k$ to \textbf{LS-C} are shown as lines. Were \cref{eq:k} a perfect fit, the point-wise estimates would follow horizontal lines: rather, point-wise $k$  changes as a function of $e_i$ (SNR), peaking at around 80-90\% errors. Generally, the fitted $k$ tends to match the point-wise estimates for $e_i\approx  0.5$. We reason that point-wise $k$ in this region are the best approximation for the global fit.
Since error rates rarely fall below 50\% on CORAAL, its $k$ values in \cref{tab:k} are likely inflated. Nonetheless, as \cref{fig:pointwise-k} illustrates, point-wise $k$ on \textbf{CL} routinely match or exceed \textbf{LS}, indicating no less an impact of textual predictability.

\section{Limitations} \label{sec:limitations}

Our profile of a small subset of ASR systems and types of noise may lead to an incomplete picture of the behaviour and utility of $k$. The choice of the LM measuring predictability may not be anodyne: mismatch between LM and ASR induced by different vocabularies (words, sub-words, characters) is not uncommon, and certainly worth exploring. 
Though $k$ fits the data well as as a single-parameter, interpretable estimate of the effects of predictability on ASR performance, as mentioned in \cref{sec:related,sec:results}, the model itself is simplistic. $k$ fails to capture the fact that increased SNRs lead to higher $k$, in both humans and ASR systems.  More complicated models of predictability involving combinatorics of errors could be fit to account for the failures of $k$ \citep[see][]{bronkhorstModelContextEffects1993}. Indeed, it could be the case that, as $e_i$ approaches $0$, so, too, do differences between ASR systems.

\section{Summary and Discussion} \label{sec:discuss}

It has been long understood that textual predictability plays an important role in ASR performance \citep{bahlMaximumLikelihoodApproach1983}, but little has been done to quantify this.
We have shown that the impact of textual predictability on ASR performance can be quantified by estimating a global ratio of log errors, $k$, across a range of acoustic conditions, showing a reliable increase in $k$ as the gap in predictability rises. 
We also see $k$ increase as a function of the ASR systems' (implicit or explicit) language modelling capacity. 
For example, as Wav2vec 2.0-Large (\textbf{W2V2-L}) \citep{baevskiWav2vecFrameworkSelfsupervised2020} has $k \approx 1.6$,  greater than the other systems tested, we conclude that it depends strongly on textual predictability for its performance.

When applied to the Corpus of Regional African-American Language \citep{kendallCorpusRegionalAfrican2023}, all systems' $k$ values increased. Though these out-of-domain data were less predictable to LMs trained on in-domain data (Librispeech, \citep{panayotovLibrispeechASRCorpus2015}), \emph{pace} \citet{martinUnderstandingRacialDisparities2020}, higher $k$ indicate that this disagreement is slight. We interpret this as supporting the notion that, for this case, improvements to ASR should focus on acoustic modelling \citep{koeneckeRacialDisparitiesAutomated2020,wassinkUnevenSuccessAutomatic2022}.


We propose  $k$ as a crucial complement to  error rates in ASR research.  We recognize that, as a general-purpose tool, the calculation of $k$ by decoding on a wide range of SNRs can be cumbersome. The following simplified recipe may be employed: \textbf{(1)} split  a corpus into high and low NLL based on an LM trained on textually similar data; \textbf{(2)} add noise until the high-NLL condition yields an error rate of around 50\%;
\textbf{(3)}  estimate $k$ as the ratio of point-wise log error rates.  The 50\% point follows from \cref{fig:pointwise-k}, but any reference rate may be used that is large enough to permit improvement in error rates.  We hope this straightforward recipe will push researchers to carefully weigh their options when choosing what aspects of ASR models are most worth improving, and which are already close to being optimal.


\newpage

\section{Acknowledgements}
\ifinterspeechfinal
This research is funded by the Data Sciences Institute at the University of Toronto and by the Natural Sciences and Engineering Research Council of Canada (NSERC) RGPIN-2022-04431. It was also enabled in part by support provided by Compute Ontario (\url{https://www.computeontario.ca/}) and the Digital Research Alliance of Canada (\url{https://alliancecan.ca/}).
\else
[to ensure author anonymity, acknowledgements will be added after the review process]
\vspace{4em}
\fi

\bibliographystyle{IEEEtranN}
\bibliography{is2024}

\end{document}